\documentclass[letterpaper, 10 pt, conference]{IEEEconf}

\usepackage{url}            % simple URL typesetting
\usepackage{booktabs}       % professional-quality tables
\usepackage{amsfonts}       % blackboard math symbols
\usepackage{nicefrac}       % compact symbols for 1/2, etc.
\usepackage{microtype}      % microtypography
\usepackage{xcolor}         % colors
\usepackage{amssymb}  % assumes amsmath package installed

\usepackage{graphicx}
\usepackage{amsmath}
\usepackage{booktabs}
\usepackage{algorithm2e}
\usepackage{setspace}
\usepackage{bbm}
\usepackage[]{caption}
\usepackage{subcaption}
\usepackage{tabularx}
\usepackage[rightcaption]{sidecap}
\usepackage{pifont}% http://ctan.org/pkg/pifont
\usepackage{xcolor}
\usepackage{pifont}
\usepackage{color, colortbl}
\usepackage{siunitx}
\usepackage{wrapfig}
\usepackage{floatrow}
\newfloatcommand{capbtabbox}{table}[][\FBwidth]

\definecolor{brightgreen}{rgb}{0.4, 1.0, 0.0}
\definecolor{LightCyan}{rgb}{0.88,1,1}
\definecolor{LightGray}{gray}{0.9}
\definecolor{Gray}{gray}{0.9}
\newcommand*\colourcheck[1]{%
  \expandafter\newcommand\csname #1check\endcsname{\textcolor{#1}{\ding{52}}}%
}
\newcommand*\colourx[1]{%
  \expandafter\newcommand\csname #1x\endcsname{\textcolor{#1}{\ding{55}}}%
}
\usepackage{amsfonts}

\colourcheck{blue}
\colourcheck{green}
\colourcheck{red}
\colourcheck{brightgreen}
\colourcheck{black}
\colourcheck{gray}
\colourx{blue}
\colourx{green}
\colourx{red}
\colourx{black}
\colourx{gray}

\usepackage{hyperref}
\hypersetup{
    colorlinks=true,
    linkcolor=orange,
    filecolor=magenta,      
    urlcolor=orange,
    citecolor=orange,
}

\usepackage[capitalise, nameinlink]{cleveref}
\newcolumntype{a}{>{\columncolor{Gray}}c}

\usepackage{multirow}

\crefname{appendixfigure}{App. Fig.}{App. Figs.}
\crefname{appendixtable}{App. Table}{App. Tables}

\usepackage{mdframed}
\usepackage[cachedir=.]{minted}
\usepackage[export]{adjustbox}

\usepackage{tcolorbox}
\usepackage{listings}
\usepackage{geometry}
\definecolor{systemprompt}{RGB}{52, 73, 94}
\definecolor{userprompt}{RGB}{41, 128, 185}
\definecolor{variable}{RGB}{192, 57, 43}
\definecolor{instruction}{RGB}{39, 174, 96}
\definecolor{example}{RGB}{142, 68, 173}
\definecolor{lightgray}{RGB}{236, 240, 241}

\tcbuselibrary{breakable}
\tcbset{
    promptbox/.style={
        breakable,
        colback=lightgray,
        colframe=black,
        boxrule=0.5pt,
        arc=2mm,
        left=5pt,
        right=5pt,
        top=5pt,
        bottom=5pt,
    }
}

\newcommand{\placeholder}[1]{\textcolor{variable}{\texttt{[#1]}}}

\newcommand{\st}{\ensuremath{s}}

\newcommand{\ac}{\ensuremath{a}}
\newcommand{\St}{\ensuremath{\mathcal{S}}}
\newcommand{\Ac}{\ensuremath{\mathcal{A}}}

\newcommand{\reward}{\ensuremath{\mathcal{R}}}

\let\labelindent\relax
\usepackage{enumitem}

\title{\LARGE \bf
Masked IRL: LLM-Guided Reward Disambiguation from Demonstrations and Language
}

\author{
  Minyoung Hwang$^1$, Alexandra Forsey-Smerek$^1$, Nathaniel Dennler$^1$, Andreea Bobu$^1$\\
  $^1$MIT CSAIL 
}

\newcommand*\maskedirllegend{\includegraphics[height=7.45pt]{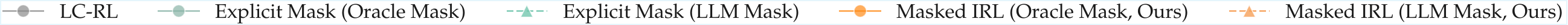}}

\begin{document}
\maketitle

%===============================================================================

\begin{abstract}
Robots can adapt to user preferences by learning reward functions from demonstrations, but with limited data, reward models often overfit to spurious correlations and fail to generalize. 
This happens because demonstrations show robots how to do a task but not what matters for that task, causing the model to focus on irrelevant state details.
Natural language can more directly specify what the robot should focus on, and, in principle, disambiguate between many reward functions consistent with the demonstrations. However, existing language-conditioned reward learning methods typically treat instructions as simple conditioning signals, without fully exploiting their potential to resolve ambiguity. Moreover, real instructions are often ambiguous themselves, so naive conditioning is unreliable. Our key insight is that these two input types carry complementary information: demonstrations show \textit{how} to act, while language specifies \textit{what} is important. We propose \emph{Masked Inverse Reinforcement Learning (Masked IRL)}, a framework that uses large language models (LLMs) to combine the strengths of both input types. Masked IRL infers state-relevance masks from language instructions and enforces invariance to irrelevant state components. When instructions are ambiguous, it uses LLM reasoning to clarify them in the context of the demonstrations. In simulation and on a real robot, Masked IRL outperforms prior language-conditioned IRL methods by up to 15\% while using up to 4.7 times less data, demonstrating improved sample-efficiency, generalization, and robustness to ambiguous language.\\
\href{https://MIT-CLEAR-Lab.github.io/Masked-IRL}{Project page} and
Code: \href{https://github.com/MIT-CLEAR-Lab/Masked-IRL}{https://github.com/MIT-CLEAR-Lab/Masked-IRL}
\end{abstract}
%===============================================================================

\section{Introduction}
\begin{figure}[t!]{
\centering
  \begin{center}
\includegraphics[width=1\textwidth]{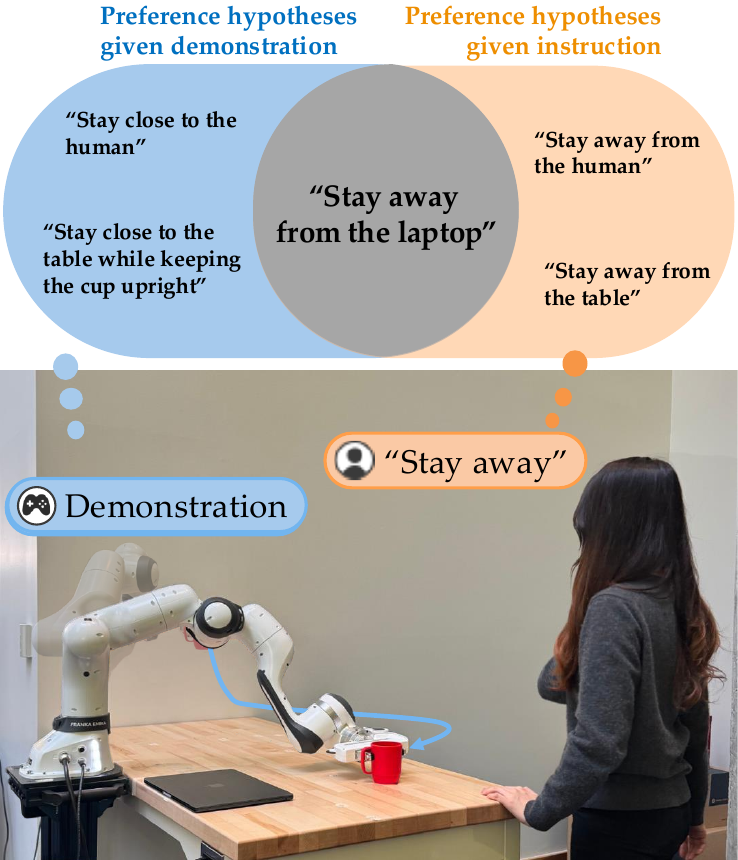}
  \end{center}
  \vspace{-10pt}
  \caption{
  \small{\textbf{Overview.} 
  Demonstrations show how to complete a task, but the same demonstration can be supported by many reward hypotheses. Language can be leveraged to disambiguate what matters in the environment. Even when both the demonstration (blue trajectory) and the associated instruction (e.g., “Stay away”) are individually ambiguous, when reasoning jointly about the pair they can often disambiguate each other, revealing the intended preference (“Stay away from the laptop”).
  }\vspace{-5mm}
  }\label{fig:overview}
  }
\end{figure}

Robots can learn how to do tasks for people by learning reward functions from user demonstrations, but reward learning is fundamentally ill-posed: many reward functions can explain the same demonstration. For instance, in the example in \cref{fig:overview}, from the demonstration alone the robot could infer it should prioritize staying close to the human, avoiding the laptop, both, or something else entirely, like producing curved trajectories. While more data could help resolve this ambiguity, in practice demonstrations are costly and difficult to collect in sufficient diversity. As such, reward models often overfit, latching onto \emph{spurious correlations} in the demonstrations rather than capturing true user intent~\cite{bobu2022inducing}.

The core issue is that, while demonstrations show robots how to perform a task, they don’t explicitly convey what matters for the task. Natural language (e.g., ``Stay away from my laptop'') could address this challenge by directly specifying what the robot should focus on and, in principle, help disambiguate between the many reward functions consistent with the demonstrations. However, existing language-conditioned reward learning methods treat language utterances as simple conditioning signals for multitask learning~\cite{fu2019language}, without fully exploiting their potential to resolve ambiguity. Moreover, real instructions are often underspecified or ambiguous: if the user in \cref{fig:overview} simply says ``Stay away'', the robot cannot determine whether to avoid the laptop, table, or human. In summary, both demonstrations and language alone are insufficient for reliable reward learning.

Our key insight is that these two input types are complementary: demonstrations show \textit{how} to act, while language specifies \textit{what} is important. From the example shown in \cref{fig:overview}, if the robot reasoned jointly about the demonstration and the instruction, it could infer that the human meant to stay away from the laptop, and thus learn the intended preference. To enable this kind of joint reasoning, we need methods that can both extract what matters from language and clarify ambiguous instructions in the context of demonstrations. 

We introduce \emph{Masked Inverse Reinforcement Learning (Masked IRL)}, a multitask reward learning framework that integrates demonstrations and natural language instructions to overcome the limitations of existing language-conditioned reward learning. Whereas prior approaches use language solely to condition rewards across multiple preferences, Masked IRL additionally exploits language to resolve ambiguity when instructions are underspecified. Specifically, our method uses LLMs in two ways: (i) to infer state-relevance masks from language instructions, enabling a masking loss that enforces invariance to irrelevant state components and reduces spurious correlations; and (ii) to clarify ambiguous instructions by using information from demonstrations, allowing reward models to remain reliable even when language is underspecified. In both simulation and real-robot experiments with a 7DoF arm, we show that combining these complementary forms of human feedback enables our method to recover more generalizable rewards  while requiring up to 4.7× fewer demonstrations 
%with far less input 
than prior language-conditioned approaches. 

In summary, our contributions are:
\textbf{(1)} Introducing language-guided state relevance masks and a novel masking loss that improves sample efficiency in IRL,
\textbf{(2)} Developing an LLM-based disambiguation mechanism that clarifies underspecified instructions using demonstrations, and
\textbf{(3)} Demonstrating robust generalization in both simulation and real-robot experiments.

\section{Related Work}

\smallskip
\noindent\textbf{Reward Learning from Human Feedback.}
An effective approach for learning robot tasks is inferring a policy or reward function from human inputs like demonstrations~\cite{ziebart2008maximum}, corrections~\cite{bajcsy2017phri}, teleoperation~\cite{zurek2021casa}, comparisons~\cite{christiano2017deep},  or trajectory rankings~\cite{brown2020brex}, among others.
To learn in a tractable way from such human data, classical IRL methods rely on hand-specified feature functions~\cite{ziebart2008maximum}, but poorly chosen features risk misalignment with human intent that produces unsafe behavior~\cite{lourenco2023diagnosing}. Deep IRL methods mitigate this assumption by learning directly from raw state, but require a large numbers of demonstrations to avoid overfitting to spurious correlations in irrelevant state components~\cite{bobu2022inducing}. These challenges make standard IRL impractical for settings where robots must adapt to diverse user preferences.

To reduce the human burden of collecting demonstrations, reinforcement learning from human feedback (RLHF) uses pairwise trajectory comparisons~\cite{christiano2017deep}. While these labels are easier for humans to provide than demonstrations, they contain at most one bit of information about the human's internal reward. Thus, RLHF often requires thousands of feedback queries to learn a single reward function~\cite{hwang2023sequential}. Recent work explores leveraging API-based LLMs to generate reward functions directly~\cite{Yu2023, hwang2024promptable}, for example by translating high-level instructions into dense rewards that can be optimized with RL~\cite{Yu2023}. Other approaches focus on personalization, either by separating feature learning from reward learning~\cite{bobu2023sirl,bobu2022inducing} or by modeling latent user preferences from feedback~\cite{Yang2024}. 

Although these methods improve efficiency for personalization, they still require training a new reward function for each user preference that do not generalize to unseen instructions. In contrast, our work learns a \emph{single} language-conditioned reward model that generalizes across preferences by using language as structured supervision.

\smallskip
\noindent\textbf{Language-Conditioned Learning in Robotics.}
Language provides a natural interface for specifying goals, feedback, and constraints in robot learning, and recent work has explored conditioning policies and rewards on natural language~\cite{fu2019language,Cui2023}. Fu et al.~\cite{fu2019language} introduced a language-conditioned reward learning approach, grounding instructions through IRL to improve transfer to novel tasks. Systems such as LILAC~\cite{Cui2023} allow operators to provide online language corrections during task execution, demonstrating how language can adapt behavior in real time. However, these methods assume instructions are clear and unambiguous, which limits their adaptability 
when language is vague or context-dependent.

% With the advent of LLMs, language has also been used for high-level planning, program synthesis, and reward specification. SayCan~\cite{Ahn2022} grounds instructions with an LLM and constrains execution using a value function, enabling robots to follow abstract commands. Other approaches treat LLMs as zero-shot planners~\cite{Huang2022b} or integrate environment feedback into planning, as in \cite{Huang2022}. Code-as-Policies~\cite{Liang2023} extends this idea by prompting LLMs to generate executable code, yielding interpretable and adaptable robotic skills. In parallel, several works explore language-conditioned reward learning with LLMs. \cite{Yu2023} translates instructions into parametric reward functions, bridging natural language and reinforcement learning. \cite{Karamcheti2023} aligns video and text for language-driven imitation learning, while \cite{hwang2025motif} learns a success detector that grounds robot motions in language semantics. Together, these approaches highlight the promise of LLMs for connecting language and control, but they still treat instructions as static inputs, rather than reasoning about which state elements matter or how to resolve underspecified commands. 

While LLMs have recently bridged language and robotic control through high-level planning \cite{Ahn2022, Liang2023} and reward specification \cite{Yu2023, hwang2025motif}, existing approaches typically treat instructions as static inputs. Consequently, these frameworks often fail to reason about state relevance or dynamically resolve underspecified commands.

In contrast to existing language-conditioned learning methods, our work uses LLMs not only to condition a shared reward model, but also to structure learning by generating state relevance masks and clarifying ambiguous instructions. This enables us to use language both as a conditioning signal and as a supervisory cue for which state elements matter, reducing data requirements.

% \smallskip
% \noindent\textbf{Abstractions in Robot Learning.}
% Many robotic tasks involve high-dimensional sensory inputs where only a subset of state components are relevant for determining the reward~\cite{bobu2024aligning}. Prior work has explored contextual feature selection to address this challenge~\cite{Ghosal2023,forsey2025context}. ~\cite{Ghosal2023} uses conditional gating mechanisms to modulate feature importance based on task context, reducing dimensionality and improving robustness. Peng et al.~\cite{peng2024adaptive} iteratively generate and validate semantically meaningful features using language-guided explanations, while Peng et al.~\cite{peng2024learning,peng2024preference} leverage background knowledge from language models to build state abstractions for unseen tasks. These approaches highlight the value of adaptively 
% identifying relevant features, but they rely on auxiliary context variables or offline feature engineering. In contrast, our work uses LLMs to infer state relevance masks directly from demonstrations and instructions, integrating feature learning into reward learning in a way that is both task-aware and data-efficient.
% \vspace{-3pt}

\section{Problem Formulation}
\label{sec:problem-formulation}
Our goal is to learn a single reward function that captures diverse human task preferences from a minimal amount of language-labeled demonstrations.

\smallskip
\noindent\textbf{Preliminaries.}
We build on the IRL framework where a human's task preference is represented as a reward function in a Markov Decision Process (MDP)~\cite{puterman2014markov} $\mathcal{M} = \langle \St, \Ac, \mathcal{T}, r \rangle$ with states $\st\in\St$, actions $\ac\in\Ac$, transition probability $\mathcal{T}:\St \times \Ac \times \St \rightarrow [0,1]$, and rewards $r:\St \rightarrow \mathbb{R}$. A solution to the MDP is a policy $\pi:\mathcal{S} \rightarrow \mathcal{A}$ that maximizes the reward and specifies what actions the robot should take in every state. The robot executes trajectories $\tau = \{s^0, \ldots, s^T\}$ according to the policy.

Since the human's reward function is not known \textit{a priori}, IRL attempts to learn it from data. In realistic settings, robots must handle \textit{many different user preferences}, each corresponding to a different underlying reward. Training a separate reward model for each preference requires extensive data, motivating a multitask formulation where a single model can generalize across preferences.

\smallskip
\noindent\textbf{Language-Conditioned Reward Learning (LC-RL).}
Language offers a natural interface for multitask reward learning by conditioning the reward on user preferences in the form of language commands~\cite{fu2019language}. 
Specifically, we consider the setting where a robot must learn a language-conditioned reward function that captures a set of human preferences $\mathcal{P} = \{P_1, \dots, P_N\}$. For each preference $P_i \in \mathcal{P}$, the human gives a set $\mathcal{D}_i = \{ (\tau^{k}_i, \ell^{k}_i) \}_{k=1}^{M_i}$ of \emph{language-labeled demonstrations}. The overall training dataset is then $\mathcal{D} = \bigcup_{i=1}^N \mathcal{D}_i$.

We parameterize the reward as a language-conditioned function $r_\theta(s \mid \ell)$, and aim to learn $\theta$ from demonstration-language pairs. The robot can then perform the task according to the preference represented by language command $\ell$ by selecting a trajectory $\tau$ that maximizes the cumulative reward $\mathcal{R}_\theta(\tau \mid \ell) = \sum_{s \in \tau} r_\theta(s\mid\ell)$.

Using language-labeled demonstrations $\mathcal{D}$, the robot infers reward parameters $\theta$ that define the human's underlying objective function. 
Inspired by prior work on language-conditioned reward learning~\cite{fu2019language}, we train our reward model using the standard Maximum Entropy IRL objective ~\cite{ziebart2008maximum}.
We model the human as a noisily rational agent who selects trajectories with probability proportional to their exponentiated reward: 
\begin{equation}
    p(\tau \mid \ell, \theta) = \frac{e^{\mathcal{R}_\theta(\tau \mid \ell)}}{
        \int_{\bar{\tau}}e^{\mathcal{R}_{\theta}(\bar{\tau} \mid \ell)} d\bar{\tau}}\enspace \propto 
    \mathrm{exp} (\reward_\theta(\tau \mid \ell)) \enspace,
    \label{eq:traj-prob}
\end{equation}
where $\ell$ captures the human's personal preference.
To recover the reward parameters, we minimize the negative log-likelihood of the demonstrations via gradient descent:
{\small
\begin{equation}
    \theta^* = \arg\min_\theta \mathcal{L}_{\rm IRL}(\theta) = \arg\min_\theta \Bigl(-\sum_{\tau, \ell \in \mathcal{D}} \log p(\tau \mid \ell, \theta)\Bigr)\enspace.
    \label{eq:reward}
\end{equation}
}
To optimize this objective, we approximate the intractable integral in Eq.~\eqref{eq:traj-prob} using importance sampling as in prior work~\cite{bobu2022inducing}.

\smallskip
\noindent\textbf{Limitations of LC-RL.}
While LC-RL provides a principled framework for inferring rewards from language-demonstration pairs, LC-RL requires high sample complexity \cite{fu2019language} and often leads to spurious correlations and overfitting in low-data regimes. Furthermore, the inherent ambiguity of natural language makes it an unreliable standalone signal for capturing precise human intent.

In this work we address these challenges by leveraging two complementary properties of language and demonstrations to improve sample efficiency and avoid spurious correlations. First, language plays a dual role in reward learning: it specifies not only \textit{what task the human wants the robot to do}, enabling a single reward model to generalize across tasks, but also implicitly indicates \textit{which aspects of the environment matter for the task}, providing a signal for filtering out irrelevant state components and improving sample efficiency. 
Second, when language commands are ambiguous, examining language in the context of demonstrations can ground instructions and resolve ambiguity. 

\begin{figure*}[t!]{
\centering
  \begin{center}
\includegraphics[width=1\textwidth]{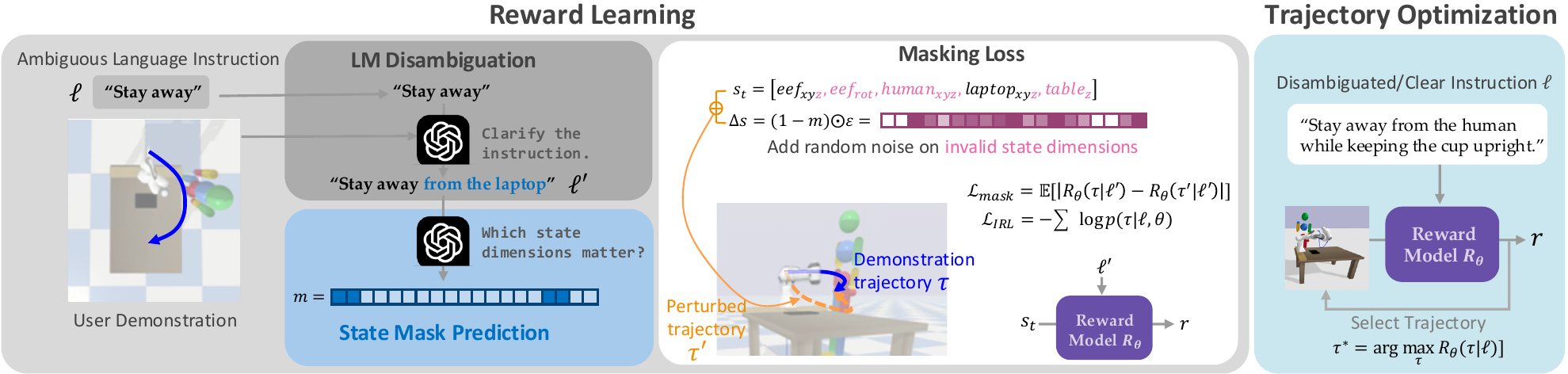}
  \end{center}
  \vspace{-10pt}
  \caption{\small\textbf{System Overview.} We clarify ambiguous language instructions using demonstrations and LLM reasoning. We then map disambiguated instructions into state masks, which guide the reward model through a masking loss that enforces invariance to irrelevant state dimensions during training. We train the reward model with the weighted sum of the masking loss and the IRL loss. Using the learned reward model, we can perform trajectory optimization by selecting the trajectory with the highest reward.\vspace{-5mm}
  }
  \label{fig:network-architecture}}
\end{figure*}

\section{Method}
We present \textbf{Masked Inverse Reinforcement Learning (Masked IRL)}, a method that leverages demonstrations paired with language instructions to efficiently learn a language-conditioned reward function. Our key contribution is to exploit common-sense priors encoded in LLMs in two complementary ways: (1) to generate \emph{relevance masks} from demonstrations paired with language (Sec.~\ref{sec:method-mask-gen}), which define a masking loss that enforces invariance to irrelevant state components (Sec.~\ref{sec:method-language-masking}); and (2) to enable training on \emph{ambiguous instructions} by reasoning about demonstrations and language in context, enabling robust reward learning even when language underspecifies the preference (Sec.~\ref{sec:method-ambiguous-lang}). The underlying structure of our approach is a language-conditioned reward model that captures shared structure across multiple preferences (Sec.~\ref{sec:method-language-conditioning}). Masked IRL remains robust under limited feedback and leverages LLM reasoning to resolve ambiguity in natural language commands. \cref{fig:network-architecture} summarizes our Masked IRL pipeline.

\subsection{Generating State Masks from Language}\label{sec:method-mask-gen}
While language conditioning provides a shared reward model across preferences, it does not by itself prevent the model from exploiting spurious correlations in irrelevant state components. To address this, we leverage LLMs to generate \emph{state relevance masks} that indicate which elements of the state vector $s$ are relevant to the instruction in context. 
For each demonstration-language pair $\{\tau, \ell\} \in \mathcal{D}$ we query an LLM with both the command $\ell$ and a description of the robot and environment state, asking it to identify which state components matter for satisfying the instruction. The LLM outputs a binary mask $m \!\in\! \{0,1\}^d$, indicating which components of the $d$ state dimensions are relevant to the instruction, where the $j^{th}$ mask element $m^{(j)}\!=\!1$ if component $j$ is relevant, and $m^{(j)}\!=\!0$ otherwise. For example, given the instruction ``Stay away from the laptop'' and a demonstration showing the robot trajectory moving around the laptop, an accurate output masks all state elements except the end effector and laptop positions (state mask prediction module in \cref{fig:network-architecture}).
We augment our training data with these LLM-generated masks, yielding $ \mathcal{D}' = \{(\tau_i, \ell_i, m_i)\}_{i=1}^M$.  We use GPT-4o for state mask prediction (see full prompts in the~\cref{sec:appendix}).

\subsection{State Masking Loss}
\label{sec:method-language-masking}
A naive way to use the state masks is \textit{explicit masking}, where we set irrelevant dimensions (corresponding to $m^{(j)}=0$) to zero. However, this approach makes the model highly sensitive to errors in mask generation, completely discarding useful state input if a state element is incorrectly masked out. Instead, we \textit{implicitly mask} irrelevant components by enforcing invariance through a masking loss, allowing the reward function to learn to ignore them without hard deletion. Formally, let $s^{(j)}$ denote a perturbed version of state $s \in \tau$, where only element $j$ is modified by adding random noise $\varepsilon$.
While various noise distributions can be used (e.g., Gaussian, uniform), we use $\varepsilon \sim \text{Uniform}(0,1)$ in our experiments. We define the masking loss $\mathcal{L}_{\rm mask}(\theta)$: 
\begin{align*}
\mathbb{E}_{\tau, \ell, m \in \mathcal{D}'}\sum_{s \in \tau}\sum_{j=1}^d
\bigl(1-m^{(j)}\bigr) \Bigl| r_\theta(s^{(j)} \mid \ell) - r_\theta(s \mid \ell)\Bigr|,
\end{align*}
which penalizes changes in the reward when irrelevant components 
are perturbed. 
The full training objective becomes a combination of the original LC-RL loss function and the masking loss,
\begin{equation}
\mathcal{J}(\theta) = \mathcal{L}_{\rm IRL}(\theta) + \lambda\, \mathcal{L}_{\rm mask}(\theta),
\end{equation}
where $\lambda>0$ is a hyperparameter controlling the trade-off between fitting the demonstrations and enforcing invariance to irrelevant state elements. We empirically compare the proposed implicit masking to naive explicit masking in Sec.~\ref{sec:experiments}.

\subsection{Clarifying Ambiguous Language Instructions}\label{sec:method-ambiguous-lang}
Natural language commands are often underspecified (e.g., ``Stay away'' without specifying to what), creating ambiguity for reward learning. We leverage LLM reasoning abilities to jointly consider language and demonstrations and hypothesize possible disambiguations. 

Following Peng et al.~\cite{peng2024adaptive}, who showed that contrasting human demonstrations with nominal robot behavior helps recover intent, we provide the LLM with: (i) a task and environment description, (ii) the language utterance $\ell$, (iii) a state-based representation of the demonstration $\tau$, and (iv) a state-based representation of the shortest-path trajectory between the same start and end points, which we call the \textit{reference trajectory}. We prompt the LLM to infer clarified commands that explain the difference between the demonstration and the reference trajectory. For example, given the instruction ``Stay away'' and a demonstration where the robot moves away from the table, the LLM may be able to reason the missing referent is the table, producing the disambiguated instruction, ``Stay away from the table.'' 

When multiple clarifications are possible (e.g., the command is ``Stay away'' and the demonstration avoids many objects), we instruct the LLM to return all disambiguations. This serves as a form of data augmentation, generating more demonstration-language pairings.

Finally, we generate state relevance masks from disambiguated commands using the same procedure as in Sec.~\ref{sec:method-language-masking}. We can now train the model conditioning on disambiguated instructions rather than the original ambiguous ones. We use GPT-5 for language disambiguation (see full prompts in the \cref{sec:appendix}).

\subsection{Language-Conditioned Reward Model Architecture}
\label{sec:method-language-conditioning}
The backbone of Masked IRL is a language-conditioned reward model with an inductive bias for sample-efficient conditioning. We encode input natural language $\ell$ with a pretrained T5 transformer~\cite{raffael2020T5} into an embedding $h_\text{lang}$. We incorporate $h_\text{lang}$ with the input state $s$ via Feature-wise Linear Modulation (FiLM)~\cite{perez2018film}. Specifically, $h_\text{lang}$ is mapped through two MLPs to produce scaling and shifting parameters $\gamma, \beta \in \mathbb{R}^d$, which transform the state input $s$: $h_\text{fused} = \gamma \odot s + \beta$. This allows the instruction to directly modulate how reward components  are computed. Compared to simple concatenation of state and language inputs, FiLM provides a more structured and efficient interface for conditioning. 
The fused representation $h_\text{fused}$ is then passed through a four-layer MLP, which maps the modulated state to a scalar reward value. We freeze the pretrained language encoder during training the reward model.

\section{Experiments}\label{sec:experiments}
We aim to evaluate the efficacy of Masked IRL to learn from limited and potentially ambiguous language and demonstrations. Our investigation seeks to answer the following research questions:
\begin{enumerate}[label=\textbf{RQ\arabic*.}, leftmargin=*, align=left]
    \item Does the proposed masking loss allow Masked IRL to efficiently learn human preferences from language and demonstrations?
    \item Do demonstrations allow us to effectively disambiguate underspecified or ambiguous language?
    \item Do our findings replicate on a physical robot interacting with a human?
\end{enumerate}

\smallskip
\noindent{\bf{Environments.}} We evaluate our research questions on an object handover task using a Franka Emika robot arm in simulation and the real world. The goal is to deliver a coffee mug from a start location to a goal location in an environment that includes a table, a laptop, and a human. The state is a 19-dimensional vector consisting of the position and rotation of the robot's end effector, objects (table and laptop), and a human in the environment. Depending on the human preference, only a subset of these state components is relevant for the reward function. This setup provides a realistic scenario where preferences can be naturally expressed in language (e.g., ``Stay away from the laptop") and grounded in demonstrations.

\begin{figure*}[t!]{\centering
    \subfloat[\centering Average Win Rate on Train Preferences after Pretraining]{
    \hspace{-10pt}
            \includegraphics[width=0.495\linewidth]{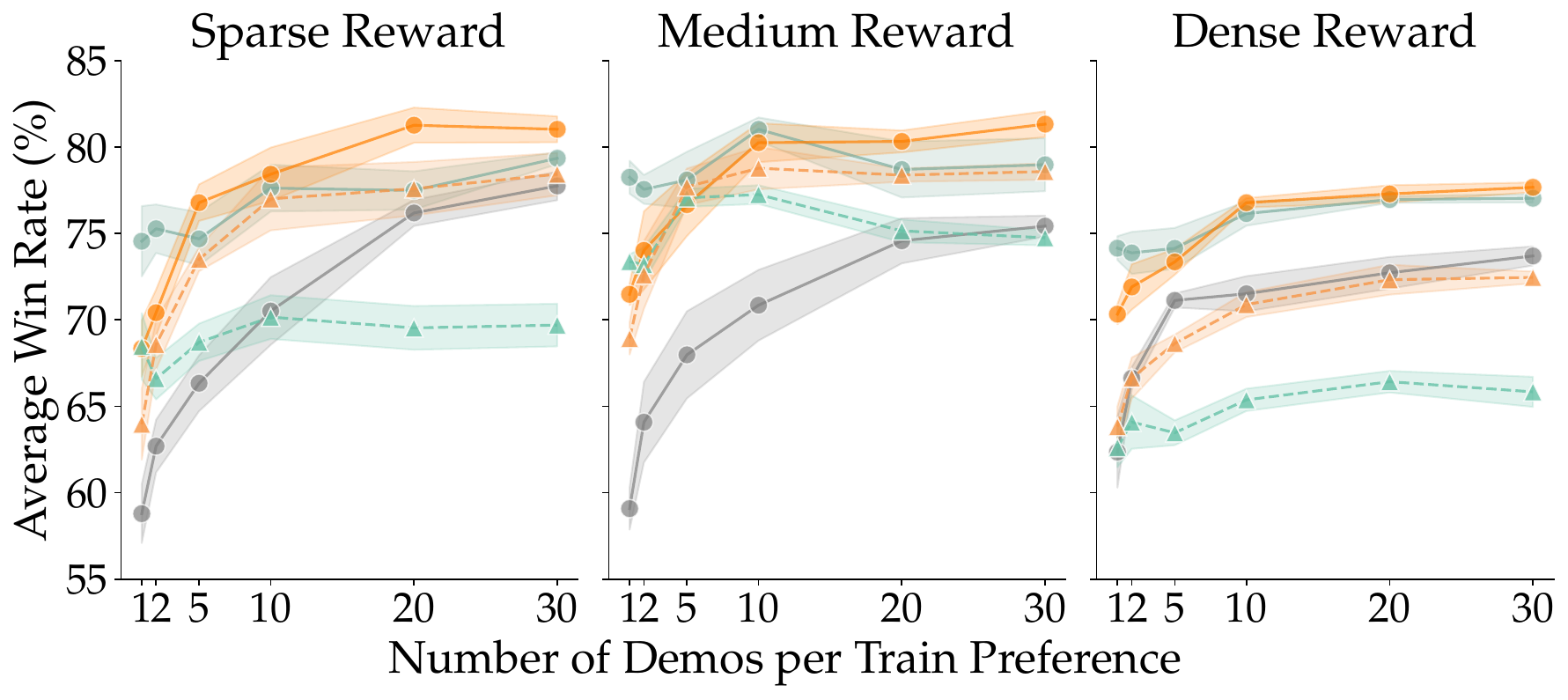}
            }\centering
    \subfloat[\centering Average Win Rate on Test Preferences after Fine-Tuning]{
            \includegraphics[width=0.495\linewidth]{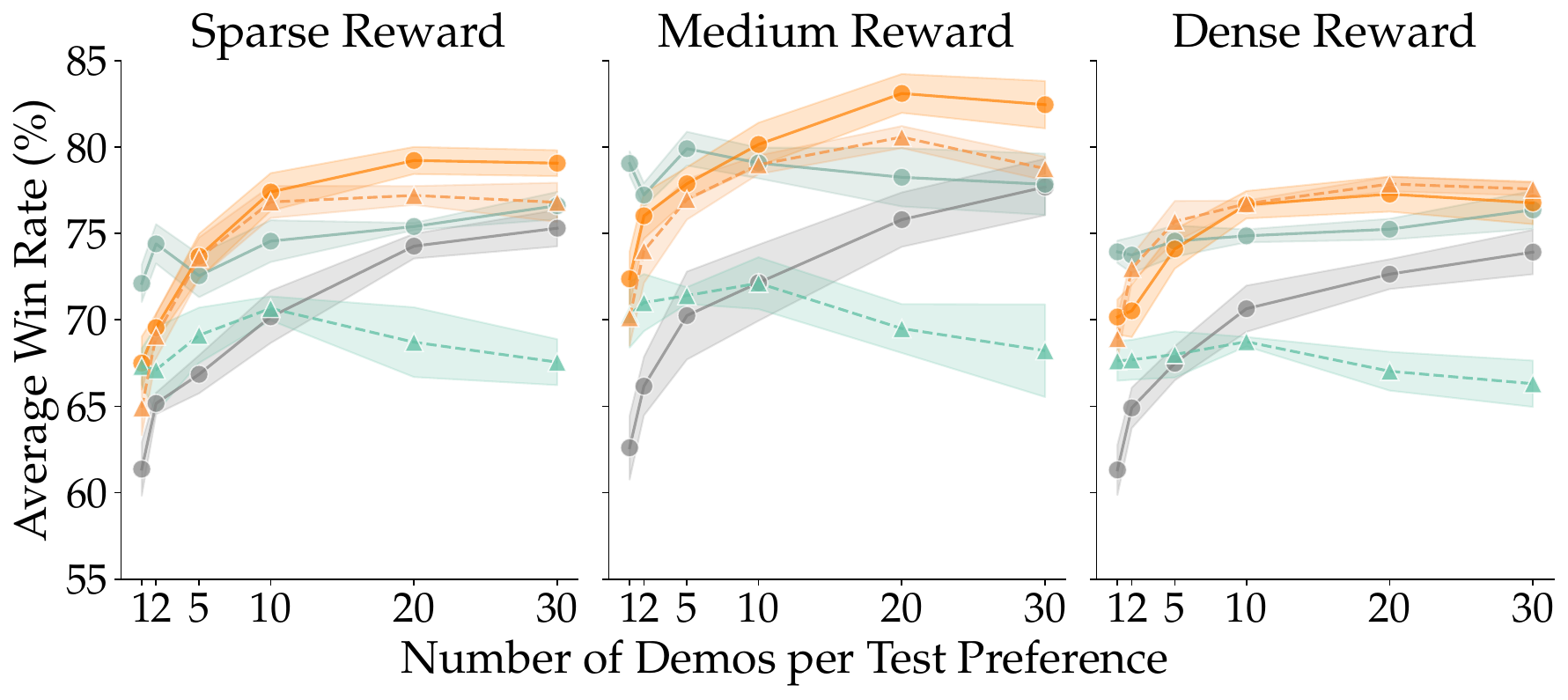}
            }\centering
    \\
    \maskedirllegend{}\vspace{-8pt}
    \caption{\small{\textbf{Performance Across Reward Densities.} 
    The average win rate of across all methods for different reward densities after (a) pretraining on 40 train preferences for 1k epochs and (b) fine-tuning on 30 test preferences for 100 epochs. All models are trained with 10 demonstrations per user preference and evaluated with unseen trajectories with novel object configurations. The shaded region indicates standard error across five different seeds.\\ % (12345, 23451, 34512, 45123, and 51234).}
    \vspace{-25pt}
    }}
    \label{fig:experiments}
}
\end{figure*}

\subsection{RQ1: Efficiency of the Masking Loss}\label{sec:rq1}

\noindent\textbf{Experimental procedure.}
Our simulated experiment is conducted in the PyBullet Simulator (\cref{fig:network-architecture}). We simulate human reward functions based on five semantic features of the robot's trajectory: distance from the table, distance from the human, distance from the laptop, distance from the human’s face, and mug orientation.
Each ground truth preference is represented by a weight vector,
where each feature is assigned a positive unit weight, a negative unit weight, or marked as irrelevant.
Positive weights indicate a preference for proximity 
(e.g., keeping the mug close to the table), negative weights indicate avoidance (e.g., staying away from the laptop), and zero weights indicate irrelevance. This formulation yields $242$ distinct preferences. We sample from this set of preferences for training and evaluating learned reward models.

We generate a trajectory dataset to train our reward models by sampling 20 object configurations and 10 start-goal pairs per configuration. For each start-goal pair, we generate 5 robot trajectories by smoothly perturbing shortest path trajectories with random noise in joint space. 
Each trajectory is paired with a language instruction that corresponds to a subset of the five features that describe the ground truth reward. 

\smallskip
\noindent\textbf{Baselines.} To answer \textbf{RQ1}, we are interested in evaluating the effectiveness of the masking loss to learn reward functions. We consider three variants of incorporating language into the loss used to train our reward networks: \textbf{(1)} \textit{Implicit Mask}, the loss proposed as in Sec.~\ref{sec:method-language-masking}; \textbf{(2)} \textit{Explicit Mask}, directly zeroing out the state dimension with the mask ($s \odot m$); and \textbf{(3)} \textit{None}, where the only language information the reward model receives is via the FiLM conditioning layer, as in LC-RL. 

Because LLMs may incorrectly infer the state mask, we additionally ablate on how the mask is generated at two variations: \textbf{(1)} \textit{LLM-generated}, the mask generated by the LLM as described in Sec.~\ref{sec:method-mask-gen}; and (2) \textit{Oracle}, a ground truth mask determined by the ground-truth human preference.
Combining all valid variants, we evaluate across the following five algorithms:
\textbf{(1)} \textit{Masked IRL (Oracle Mask)}, \textbf{(2)} \textit{Masked IRL (LLM Mask)}, \textbf{(3)} \textit{Explicit Mask (Oracle Mask)}, \textbf{(4)} \textit{Explicit Mask (LLM Mask)}, and \textbf{(5)} \textit{LC-RL}~\cite{fu2019language}.

Each variant uses the model architecture in Sec.~\ref{sec:method-language-conditioning}. 

\smallskip
\noindent\textbf{Evaluation metrics.} We evaluate our approach using the \textit{average win rate}: 
given two trajectories sampled from the test set, the learned reward predicts which one is preferred, and we score agreement with the ground-truth reward.
This measures how often the learned reward model correctly prefers better trajectories compared to ground-truth preferences. We measure the average win rate on three different reward densities: \textit{sparse}, \textit{medium}, and \textit{dense}. The density of the ground truth reward model is defined based on the number of nonzero preference weights a simulated human has for the five semantic features (sparse: 1, 2, medium: 3, dense: 4, 5). We run all experiments with 5 different random seeds and show the average and standard error across seeds.

\smallskip
\noindent\textbf{Results.} 
We first evaluate our proposed masking loss function against naive language conditioning on input layers. \cref{fig:experiments} shows that Masked IRL with both Oracle and LLM-generated masks consistently matches or outperforms the language-conditioned baseline (LC-RL) across different reward densities, both for train and test preferences. This demonstrates that naively conditioning the reward model on language is insufficient because the model can easily overfit to spurious correlations. In contrast, Masked IRL’s masking loss penalizes sensitivity to irrelevant state elements. The masked loss enables the reward model to focus on task-relevant dimensions and improves both robustness and generalization.

Another key benefit of Masked IRL is its improved sample efficiency. As shown in \cref{fig:experiments}, Masked IRL has a larger area under the win rate curve as the number of demonstrations increases.
Because the masking loss discourages dependence on irrelevant state dimensions, the model can extract more useful information from fewer demonstrations. In practice, this means Masked IRL achieves strong generalization even with as few as five demonstrations per preference, while the other baselines require substantially more data --- up to $33\%$ more for Explicit Mask and $4.7$ times more for LC-RL on average --- to reach comparable performance. This efficiency is particularly valuable in robotics, where collecting demonstrations from humans is time-consuming.

Both explicit masking and Masked IRL outperform LC-RL when oracle masks are provided, but performance diverges under noisy LLM-generated masks, shown in \cref{fig:experiments}. Explicit masking with LLM masks performs poorly, especially as the number of demonstrations increases, likely because hard-masking prohibits the model learning from state components that are potentially relevant to the preference due to noise. 
In contrast, Masked IRL remains robust with LLM masks: the masking loss encourages the model to adapt to multiple preferences even with imperfect supervision, preventing collapse and yielding stable gains over LC-RL.

\subsection{RQ2: Robustness to Language Ambiguity}\label{sec:RQ2}

\noindent\textbf{Experimental Procedure.}
We use the same demonstrations as described in Sec.~\ref{sec:rq1}, but demonstrations are instead paired with \textit{ambiguous instructions} based on the ground truth preferences. We procedurally generate ambiguous instructions that deliberately underspecify the user's preference in two ways naturally done by humans~\cite{wan2025infer}: (1) \emph{referent-omitted} commands, which specify a relation without the object (e.g., ``Stay close''), and 
(2) \emph{expression-omitted} commands, which specify an object but not the relation (e.g., ``Table''). Because simultaneously omitting referents or expressions for multiple features would yield contrived and linguistically unnatural commands (e.g., ``Stay away from this and stay away from that and stay close to another one''), we restrict our evaluation to sparse rewards. In this experiment, only a single feature is active at a time, allowing us to generate ambiguous commands that are both natural and representative of how users might underspecify preferences. Specifically, we evaluate six different sparse rewards, each defined by a positive or negative weighting over one of the three features: distance to the table, distance to the laptop, and distance to the human. For each preference, we assess our disambiguation method on both referent-omitted and expression-omitted instructions, paired with 10 demonstrations.

\smallskip
\noindent\textbf{Baselines.} To answer \textbf{RQ2}, we are interested in learning user preferences when given ambiguous language inputs. We consider the same three variants of incorporating language into the loss as Sec.~\ref{sec:rq1}: \textbf{(1)} \textit{Masked IRL}, \textbf{(2)} \textit{Explicit Mask}, and \textbf{(3)} \textit{LC-RL}. Our proposed approach performs a disambiguation process described in Sec.~\ref{sec:method-ambiguous-lang}. The disambiguated instructions are used to predict state masks, following Sec.~\ref{sec:method-language-masking}. To evaluate the effectiveness of the disambiguation step we use two variants of incorporating the instruction information: \textbf{(1)} \textit{Disambiguated Instructions} (DI), the proposed disambiguation pipeline; and (2) \textit{Ambiguous Instructions} (AI), directly calculating the mask from ambiguous instructions without first disambiguating the instructions.

\smallskip
\noindent\textbf{Evaluation metrics.}
For evaluating average win rate, we provide ambiguous instructions to models labeled ``AI" and disambiguated instructions to models labeled ``DI". 
We further evaluate LLM disambiguation performance with two additional metrics: \textit{instruction accuracy} and mask-based \textit{Precision, Recall, and F1} scores. We define a disambiguation query as correct if the generated set of command candidates includes an instruction semantically equivalent to the ground-truth clarified command. To account for LLM stochasticity, we report the average accuracy across five independent query rounds. For training the DI reward model baselines, we select the clarified instructions from the single most accurate round.

\begin{table}[t!]
\resizebox{\textwidth}{!}{
\begin{tabular}{lccc}
\toprule
\textbf{Instruction Type} & \textbf{Precision} & \textbf{Recall} & \textbf{F1 Score} \\ \midrule
Ambiguous & 0.531 $\pm$ 0.003 & 0.910 $\pm$ 0.009 & 0.670 $\pm$ 0.005 \\
Disambiguated & 0.705 $\pm$ 0.001 & 0.882 $\pm$ 0.024 & 0.783 $\pm$ 0.009 \\
\arrayrulecolor{gray}
\midrule
\arrayrulecolor{black}
Clear & 0.789 $\pm$ 0.017 & 1.000 $\pm$ 0.000 & 0.882 $\pm$ 0.010 \\
\bottomrule
\end{tabular}}
\caption{
\textbf{State Mask Prediction from Different Instruction Types.} 
Disambiguated instructions improve all metrics over ambiguous instructions. Errors denote standard errors across five runs.
}
\label{tab:language-to-state-mask-prediction}
\end{table}

\begin{figure}[t]{
\centering
  \begin{center}
\includegraphics[width=1\textwidth]{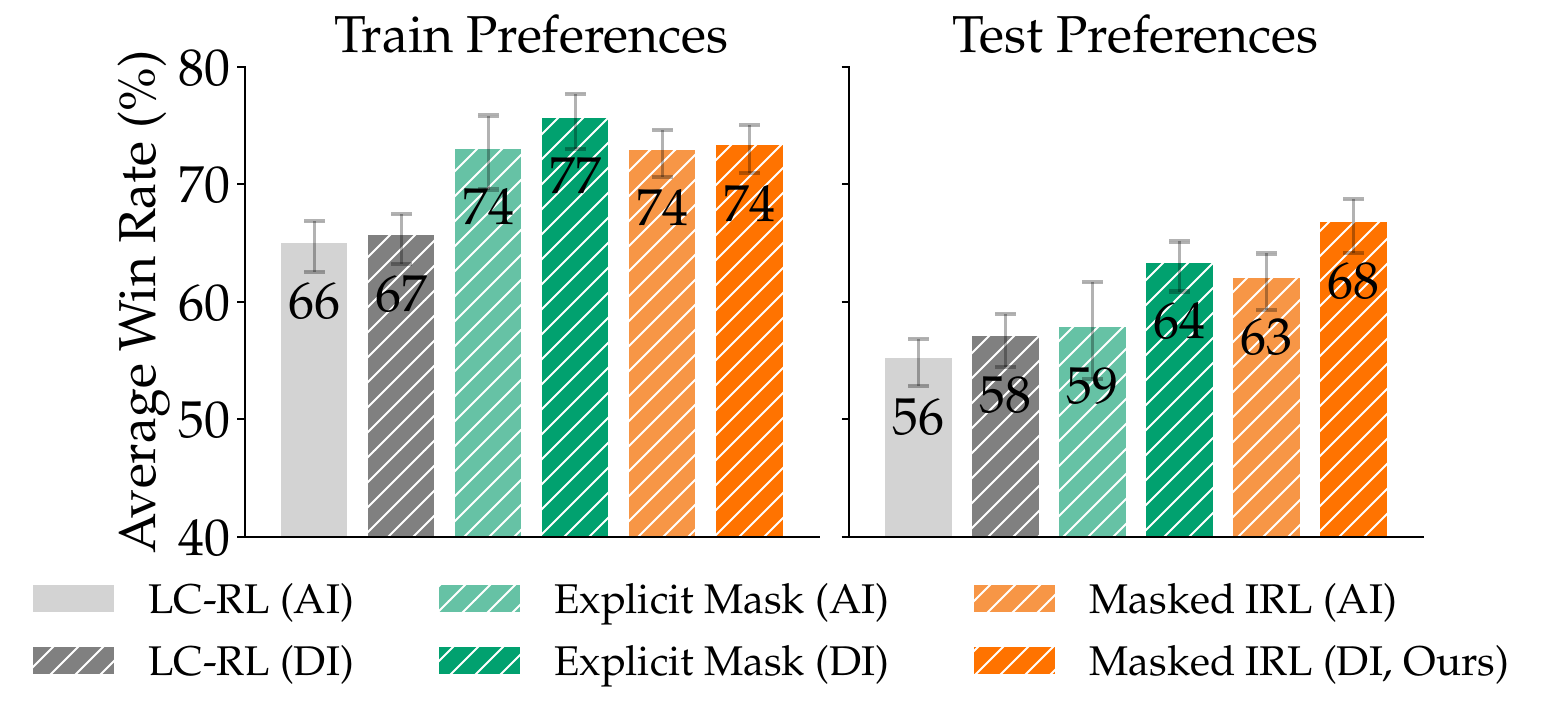}
  \end{center}
  \vspace{-10pt}
  \caption{
  \small{\textbf{Performance on ambiguous language.} AI and DI denote models trained with ambiguous and disambiguated instructions, respectively. While LC-RL only uses language to condition the reward model, both Explicit Mask and Masked IRL significantly outperform LC-RL on train preferences, demonstrating the benefit of our masking approach. On unseen test preferences, both language disambiguation and masking are important, where Masked IRL using disambiguated instructions show the highest performance.}\vspace{-10pt}
  }\label{fig:ambiguous-language}
  }
\end{figure}

\smallskip
\noindent\textbf{Results.}
Over the five rounds of 6 preferences, the average instruction accuracy of the language disambiguation step of our pipeline was $76.4\%$, and the average number of disambiguated instruction candidates per ambiguous command-demo pair was $1.12$. We also measure the state mask prediction performance from clear, ambiguous, and disambiguated instructions, as shown in \cref{tab:language-to-state-mask-prediction}. While clear instruction leads to the highest performance in all metrics, disambiguated instructions show $16.9\%$ higher F1 score than ambiguous instructions.
\cref{fig:ambiguous-language} shows the performance of reward learning using ambiguous or disambiguated instructions. On test preferences, using disambiguated language improves performance for all methods: LC-RL, Explicit Mask, and Masked IRL. Masked IRL trained with disambiguated instructions shows the highest generalization performance, showing $21.4\%$ higher average win rate than LC-RL trained with ambiguous instructions. 

\subsection{RQ3: Evaluation in the Real World}
For \textbf{RQ3}, we are interested in the efficacy of our approach in the real world, where human demonstrations may be suboptimal with respect to specified preferences. We conduct experiments on the robot using the same set of models as in Sec.~\ref{sec:rq1}: LC-RL, \{Explicit Masking, Implicit Masking\} $\times$ \{Oracle Mask, LLM Mask\}.

\smallskip
\noindent\textbf{Experimental Procedure.} For real world experiments, we collect $1,200$ demonstrations evenly distributed over $50$ preferences. Each preference comprises of two demonstrations for each of $12$ object configurations, with both demonstrations sharing the same start-goal pair, randomly sampled from nine possible locations. Two experts provided demonstrations by kinesthetically guiding the robot according to a given preference.

\smallskip
\noindent\textbf{Evaluation metrics.} In addition to average win rate, we evaluate \textit{average reward variance} and \textit{average regret} of trajectories optimized with learned rewards. We evaluate average reward variance by adding Gaussian noise sampled from $\mathcal{N}(0,1)$ to irrelevant state dimensions. This procedure is repeated five times, and the variances of the resulting rewards are averaged to obtain the final measure. To evaluate average regret, we first perform discrete optimization over the set of test trajectories to choose the most optimal trajectory with learned reward models given test preferences. We calculate the regret by calculating the difference of the ground truth rewards between the chosen trajectory and the actual optimal trajectory that maximizes the ground truth reward function.

\begin{figure}[t]{
\centering
  \begin{center}
  \hspace{-12pt}
\includegraphics[width=1.03\textwidth]{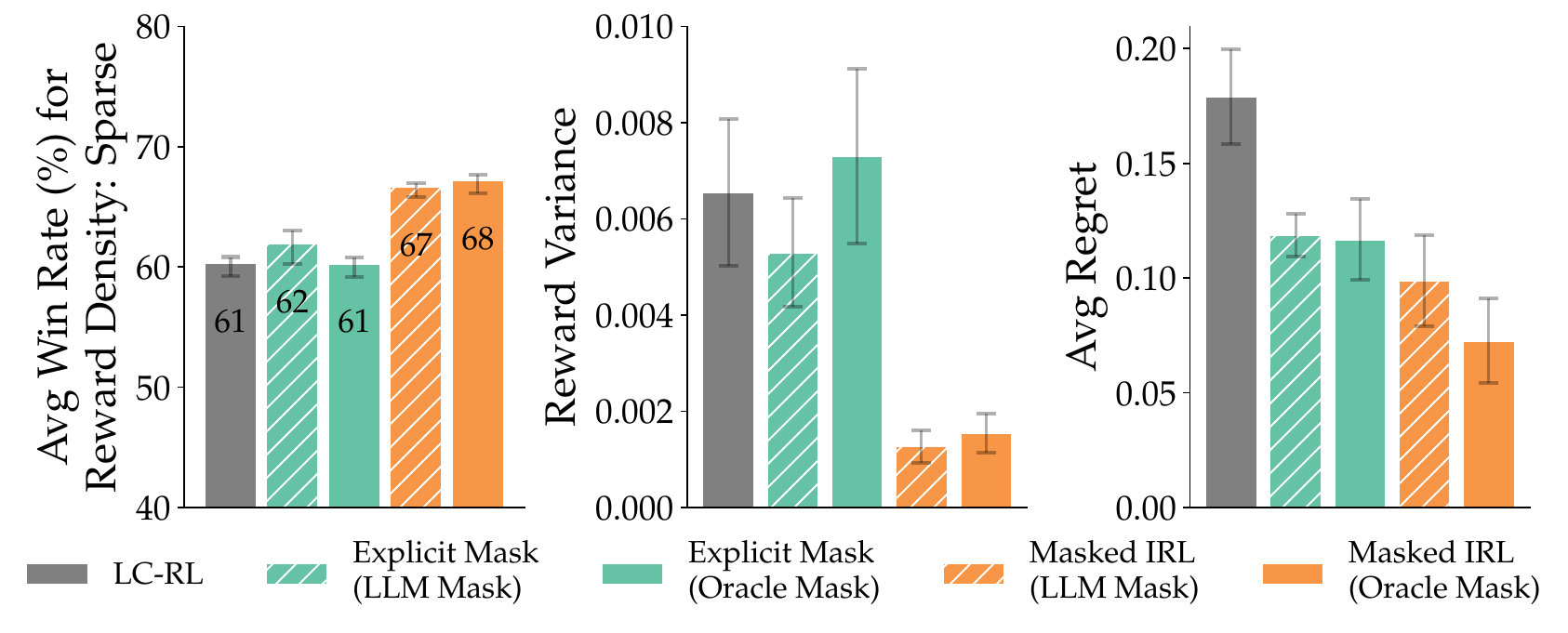}
  \end{center}
  \vspace{-10pt}
  \caption{
  \small{\textbf{Zero-shot Performance on Test Preferences with Real Robot.} Masked IRL achieves higher win rates, lower reward variance given perturbation on irrelevant state dimensions, and lower win rates on optimized trajectories than baselines, showing its effectiveness in transferring to novel preferences without additional training.}\vspace{-15pt}
  }\label{fig:realrobot}
  }
\end{figure}

\smallskip
\noindent\textbf{Zero-shot generalization to real robot.}
We further validate Masked IRL on a real Franka Panda robot. As shown in \cref{fig:realrobot}, Masked IRL achieves higher average win rates and lower reward variance than all baselines, demonstrating that this method transfers to real world human demonstrations without additional fine-tuning or architecture changes. Masked IRL additionally shows significantly lower reward variance compared to LC-RL and Explicit Mask, demonstrating that masking loss effectively enforces invariance to irrelevant state changes. These results highlight the generalization of our approach: language-guided implicit masking makes the learned rewards more robust to distributional shifts in the real world. Furthermore, the rightmost plot in \cref{fig:realrobot} shows that Masked IRL achieves $59.4\%$ and $44.8\%$ lower average reward regret than LC-RL using oracle and LLM masks, respectively. This demonstrates that Masked IRL learns rewards that lead to better optimized trajectories than baseline approaches for reward learning.

\section{Discussion}
\noindent\textbf{Conclusion.}
Reward learning from demonstrations is often ambiguous and susceptible to overfitting, since demonstrations show how to act but not what matters. To address this, we propose Masked IRL, which leverages LLMs to generate state relevance masks and incorporates a masking loss that enforces invariance to irrelevant state dimensions. Combined with an LLM-based disambiguation of underspecified instructions, this approach improves sample efficiency, robustness, and generalization, outperforming prior language-conditioned IRL methods in both simulation and real-robot experiments with up to 4.7 times fewer demonstrations.

\smallskip
\noindent\textbf{Limitations and Future Work.} Although our Masked IRL framework effectively improves generalization and sample efficiency, several limitations remain. First, our reliance on LLMs introduces potential inaccuracies in generating relevance masks, particularly when instructions are ambiguous or nuanced, which can affect the overall robustness of the reward model. Future work could explore methods for refining mask accuracy through interactive human feedback or advanced prompting strategies. Additionally, our current evaluations focus on relatively constrained robotic tasks; extending the approach to more complex, dynamic, or multi-agent environments could further validate the generality of Masked IRL. While we focus on manipulation, the framework naturally extends to other domains where humans can provide both behavioral feedback (e.g., demonstrations, corrections) and semantic feedback (e.g., language, gaze, or gestures). Lastly, investigating ways to integrate explicit uncertainty estimation in the masking process could enhance the reliability of our approach in real-world deployments.

\section*{ACKNOWLEDGMENT}
{\footnotesize This research was supported in part by the Tata Group via the MIT Generative AI Impact Consortium (MGAIC) Award, and the Department of Defense (DoD) through the National Defense Science \& Engineering Graduate (NDSEG) Fellowship Program.}

\bibliographystyle{plain}
\bibliography{references}

@inproceedings{Cui2023,
  title={``No, to the Right'' -- Online Language Corrections for Robotic Manipulation via Shared Autonomy},
  author={Cui, Yuchen and Karamcheti, Siddharth and others},
  booktitle={Proceedings of HRI 2023},
  year={2023}
}

@inproceedings{Ahn2022,
  title={Do As I Can, Not As I Say: Grounding Language in Robotic Affordances},
  author={Ahn, Michael and Brohan, Anthony and Brown, Noah and et al.},
  booktitle={Conference on Robot Learning (CoRL)},
  year={2022}
}

@inproceedings{Liang2023,
  title={Code as Policies: Language Model Programs for Embodied Control},
  author={Liang, Jacky and Huang, Wenlong and Xia, Fei and Xu, Peng and Hausman, Karol and Ichter, Brian and Florence, Pete and Zeng, Andy},
  booktitle={ICRA},
  year={2023}
}

@inproceedings{Yu2023,
  title={Language to Rewards for Robotic Skill Synthesis},
  author={Yu, Wenhao and Gileadi, Nimrod and Fu, Chuyuan and et al.},
  booktitle={CoRL},
  year={2023}
}

@inproceedings{Yang2024,
  title={Trajectory Improvement and Reward Learning from Comparative Language Feedback},
  author={Yang, Zhaojing and Jun, Miru and Tien, Jeremy and Russell, Stuart J. and Dragan, Anca D. and Bıyık, Erdem},
  journal={arXiv preprint arXiv:2410.06401},
  booktitle={Proceedings of the Conference on Robot Learning (CoRL)},
  year={2024}
}

@inproceedings{perez2018film,
  title={Film: Visual reasoning with a general conditioning layer},
  author={Perez, Ethan and Strub, Florian and De Vries, Harm and Dumoulin, Vincent and Courville, Aaron},
  booktitle={AAAI conference on artificial intelligence},
  year={2018}
}

@book{puterman2014markov,
  title={Markov decision processes: discrete stochastic dynamic programming},
  author={Puterman, Martin L},
  year={2014},
  publisher={John Wiley \& Sons}
}

@inproceedings{fu2019language,
  author       = {Justin Fu and
                  Anoop Korattikara and
                  Sergey Levine and
                  Sergio Guadarrama},
  title        = {From Language to Goals: Inverse Reinforcement Learning for Vision-Based
                  Instruction Following},
  booktitle    = {ICLR},
  year         = {2019},
}

@article{christiano2017deep,
  title={Deep reinforcement learning from human preferences},
  author={Christiano, Paul F and Leike, Jan and Brown, Tom and Martic, Miljan and Legg, Shane and Amodei, Dario},
  journal={Advances in neural information processing systems},
  year={2017}
}

@inproceedings{ziebart2008maximum,
  title={Maximum entropy inverse reinforcement learning.},
  author={Ziebart, Brian D and Maas, Andrew L and Bagnell, J Andrew and Dey, Anind K and others},
  booktitle={Aaai},
  volume={8},
  pages={1433--1438},
  year={2008},
  organization={Chicago, IL, USA}
}

@article{bobu2022inducing,
  title={Inducing structure in reward learning by learning features},
  author={Bobu, Andreea and Wiggert, Marius and Tomlin, Claire and Dragan, Anca D},
  journal={The International Journal of Robotics Research},
  volume={41},
  number={5},
  pages={497--518},
  year={2022},
  publisher={SAGE Publications Sage UK: London, England}
}

@inproceedings{bobu2023sirl,
  title={Sirl: Similarity-based implicit representation learning},
  author={Bobu, Andreea and Liu, Yi and Shah, Rohin and Brown, Daniel S and Dragan, Anca D},
  booktitle={Human-Robot Interaction},
  pages={565--574},
  year={2023}
}

@article{peng2024adaptive,
  title={Adaptive language-guided abstraction from contrastive explanations},
  author={Peng, Andi and Li, Belinda Z and Sucholutsky, Ilia and Kumar, Nishanth and Shah, Julie A and Andreas, Jacob and Bobu, Andreea},
  journal={CoRL},
  year={2024}
}

@article{hwang2025motif,
  title={MotIF: Motion Instruction Fine-tuning},
  author={Hwang, Minyoung and Hejna, Joey and Sadigh, Dorsa and Bisk, Yonatan},
  journal={Robotics and Automation Letters},
  year={2025},
  publisher={IEEE}
}

@article{hwang2023sequential,
  title={Sequential preference ranking for efficient reinforcement learning from human feedback},
  author={Hwang, Minyoung and Lee, Gunmin and Kee, Hogun and Kim, Chan Woo and Lee, Kyungjae and Oh, Songhwai},
  journal={NeurIPS},
  volume={36},
  year={2023}
}

@inproceedings{hwang2024promptable,
  title={Promptable behaviors: Personalizing multi-objective rewards from human preferences},
  author={Hwang, Minyoung and Weihs, Luca and Park, Chanwoo and Lee, Kimin and Kembhavi, Aniruddha and Ehsani, Kiana},
  booktitle={CVPR},
  year={2024}
}

@article{raffael2020T5,
  author       = {Colin Raffel and
                  Noam Shazeer and
                  Adam Roberts and
                  Katherine Lee and
                  Sharan Narang and
                  Michael Matena and
                  Yanqi Zhou and
                  Wei Li and
                  Peter J. Liu},
  title        = {Exploring the Limits of Transfer Learning with a Unified Text-to-Text
                  Transformer},
  journal      = {J. Mach. Learn. Res.},
  volume       = {21},
  pages        = {140:1--140:67},
  year         = {2020},
  url          = {https://jmlr.org/papers/v21/20-074.html},
  bibsource    = {dblp computer science bibliography, https://dblp.org}
}

@inproceedings{wan2025infer,
  title={Infer human’s intentions before following natural language instructions},
  author={Wan, Yanming and Wu, Yue and Wang, Yiping and Mao, Jiayuan and Jaques, Natasha},
  booktitle={Proceedings of the AAAI Conference on Artificial Intelligence},
  volume={39},
  number={24},
  pages={25309--25317},
  year={2025}
}

@InProceedings{bajcsy2017phri, title = {Learning Robot Objectives from Physical Human Interaction}, author = {Andrea Bajcsy and Dylan P. Losey and Marcia K. O’Malley and Anca D. Dragan}, booktitle = {Conference
on Robot Learning (CoRL)}, pages = {217--226}, year = {2017}, publisher = {PMLR}}

@InProceedings{brown2020brex, title = {Safe Imitation Learning via Fast {B}ayesian Reward Inference from Preferences}, author = {Brown, Daniel and Coleman, Russell and Srinivasan, Ravi and Niekum, Scott}, booktitle = {ICML}, year = {2020} }

@inproceedings{zurek2021casa,
  author       = {Matthew Zurek and
                  Andreea Bobu and
                  Daniel S. Brown and
                  Anca D. Dragan},
  title        = {Situational Confidence Assistance for Lifelong Shared Autonomy},
  booktitle    = {{IEEE} International Conference on Robotics and Automation, {ICRA}
                  2021, Xi'an, China, May 30 - June 5, 2021},
  pages        = {2783--2789},
  publisher    = {{IEEE}},
  year         = {2021},
  url          = {https://doi.org/10.1109/ICRA48506.2021.9561839},
  doi          = {10.1109/ICRA48506.2021.9561839},
  bibsource    = {dblp computer science bibliography, https://dblp.org}
}

@inproceedings{lourenco2023diagnosing,
  author       = {In{\^{e}}s Louren{\c{c}}o and
                  Andreea Bobu and
                  Cristian R. Rojas and
                  Bo Wahlberg},
  title        = {Diagnosing and Repairing Feature Representations Under Distribution
                  Shifts},
  booktitle    = {62nd {IEEE} Conference on Decision and Control, {CDC} 2023, Singapore,
                  December 13-15, 2023},
  pages        = {3638--3645},
  publisher    = {{IEEE}},
  year         = {2023},
  url          = {https://doi.org/10.1109/CDC49753.2023.10383644},
  doi          = {10.1109/CDC49753.2023.10383644},
  bibsource    = {dblp computer science bibliography, https://dblp.org}
}

\newpage
\clearpage
\appendix
\label{sec:appendix}
\subsection{Full Prompt}\label{sec:appendix-full-prompt}
The prompts used for language disambiguation and state mask prediction are designed to elicit consistent and interpretable reasoning from LLMs. Each prompt corresponds to a distinct reasoning phase within our framework: (1) \textit{Language Disambiguation} and (2) \textit{State Mask Prediction}. 

\textbf{Language Disambiguation Prompt.}
The goal of this prompt is to infer what aspect of the environment a human user cares about when providing a demonstration and a short language instruction.
At runtime, the placeholders \placeholder{ref\_desc}, \placeholder{demo\_desc}, and \placeholder{instruction} are filled with the specific trajectories and textual commands associated with each trial.
The system prompt describes the task and environment (a Franka Emika Panda robot carrying a coffee cup on a tabletop scene with a human and laptop) and introduces a reference ``shortest path'' trajectory that serves as a neutral baseline.
The user prompt then introduces the human demonstration and instruction, prompting the model to reason about how the demonstration differs from the shortest path.
By comparing these trajectories, the LLM identifies which visible object(s) (table, human, laptop) the demonstrator was referring to and outputs one or two disambiguated commands in JSON form, such as:
\begin{quote}
\small\texttt{["Stay close to the human"]} \quad or \quad \texttt{["Stay away from the table", \\"Stay away from the laptop"]}.
\end{quote}
This procedure transforms a potentially ambiguous instruction into a grounded set of explicit action--referent pairs.
Because each instruction is interpreted in relation to a demonstration rather than in isolation, the same phrase (e.g., ``stay away'') can acquire different meanings depending on the motion pattern the LLM observes.

\begin{tcolorbox}[promptbox, title={\textcolor{white}{\textbf{System Context}}}, colframe=systemprompt, coltitle=white, colbacktitle=systemprompt]

\textbf{Environment Description:} A Franka robot arm carries a coffee cup on a tabletop, controlled via PyBullet simulator.

\vspace{3mm}

\textbf{Trajectory Format:} The robot's trajectory is represented as a \textcolor{variable}{\textbf{19×21 matrix}}:
\begin{itemize}
    \item \textbf{21 timesteps} over the demonstration
    \item Each state is a \textbf{19-dimensional vector}:
    \begin{itemize}
        \item Positions: \textcolor{instruction}{$x, y, z$} of robot end effector (3 dims)
        \item Orientation: \textcolor{instruction}{9D rotation matrix} (9 dims)
        \item Fixed positions: \textcolor{instruction}{human $xyz$, laptop $xyz$, table $z$} (7 dims)
    \end{itemize}
\end{itemize}

\vspace{3mm}

\textbf{Reference Trajectory:} \textcolor{variable}{\texttt{[ref\_desc]}} represents the shortest path between randomly sampled start and goal points.

\vspace{3mm}

\textbf{Reasoning Instructions:}
\begin{itemize}
    \item Track robot movements \textit{relative to} important objects in the scene
    \item For distances to bulky objects (e.g., human): consider horizontal distance in the $xy$-plane
    \item Consider specific axes or planes depending on language instruction context
\end{itemize}

\end{tcolorbox}

\begin{tcolorbox}[promptbox, title={\textcolor{white}{\textbf{User Query}}}, colframe=userprompt, coltitle=white, colbacktitle=userprompt]

\textbf{Inputs Provided:}
\begin{itemize}
    \item \textbf{Demonstration:} \textcolor{variable}{\texttt{[demo\_desc]}} — user-provided trajectory (same format, same start/goal as reference)
    \item \textbf{Language Command:} \textcolor{variable}{\texttt{[instruction]}} — user's explanation of the demonstration
\end{itemize}

\vspace{3mm}

\textbf{Task:} Describe the demonstration trajectory in context of:
\begin{enumerate}
    \item The environment
    \item The language command provided
    \item Comparison to the shortest path trajectory
\end{enumerate}

\vspace{3mm}

\textbf{Goal:} Disambiguate which feature(s) the user cares about by:
\begin{itemize}
    \item Analyzing trajectory differences
    \item Reasoning about movements relative to \textit{each} object (table, laptop, human)
    \item Grounding answers in visible scene objects
\end{itemize}

\end{tcolorbox}

\begin{tcolorbox}[promptbox, title={\textcolor{white}{\textbf{Critical Constraints}}}, colframe=variable, coltitle=white, colbacktitle=variable]
\begin{enumerate}
    \item Use \textbf{EXACT wording} from original command (no paraphrasing)
    \item Each object appears in \textbf{AT MOST ONE} output command
    \item No object referenced multiple times across commands
    \item Output format: JSON list of 1–2 disambiguated commands (strings only, no extra text)
\end{enumerate}
\end{tcolorbox}

\begin{tcolorbox}[promptbox, title={\textcolor{white}{\textbf{Examples}}}, colframe=example, coltitle=white, colbacktitle=example]

\textbf{Example 1:}

\textit{Original command:} \textcolor{variable}{\texttt{"[object]"}}

$\rightarrow$ If demonstration moves near the object: 

\textcolor{instruction}{\texttt{["Stay close to the [object]"]}}

\vspace{3mm}

\textbf{Example 2:}

\textit{Original command:} \textcolor{variable}{\texttt{"Stay away."}}

$\rightarrow$ If demonstration stays farther from object than shortest path:

\textcolor{instruction}{\texttt{["Stay away from the [object]"]}}

\vspace{3mm}

\textbf{Example 3 (Axis Consideration):}

\textit{Original command:} \textcolor{variable}{\texttt{"Stay away."}}

$\rightarrow$ If trajectory goes high above table and over laptop in $xy$-plane:
\begin{itemize}
    \item Table is bulky with significant height → consider $z$-axis distance
    \item Laptop is low-profile → intention relates to $xy$-plane distance
\end{itemize}

\textcolor{instruction}{\texttt{["Stay away from the table"]}}

\vspace{3mm}

\textbf{Example 4 (Multiple Objects):}

\textit{Original command:} \textcolor{variable}{\texttt{"Stay away."}}

$\rightarrow$ If demonstration avoids two objects:

\textcolor{instruction}{\texttt{["Stay away from [object 1]", "Stay away from [object 2]"]}}

\end{tcolorbox}

\textbf{State Mask Prediction Prompt.}
The second family of prompts, shown under ``LLM Prompts Used for State Mask Prediction,'' is used to predict a binary attention mask over the robot’s 19-dimensional state representation.
Here, the LLM is asked to decide which dimensions (e.g., specific end-effector coordinates, rotation elements, or object positions) are relevant to the provided language instruction.
The context text explains the physical setup and enumerates all state variables so that the model has an explicit mapping from symbolic names to scene elements.
At inference time, the placeholder \placeholder{instruction} is replaced with the task-specific command (e.g., ``Stay close to the table surface'' or ``Keep the cup upright''), and the model outputs a structured JSON mask such as:
\begin{quote}
\small\texttt{\{"eef\_pos": [1,1,0], "eef\_rot": [0,...,1], "human": [0,0,0], "laptop": [0,0,0], "table": [1]\}}.
\end{quote}

\begin{tcolorbox}[promptbox, title={\textcolor{white}{\textbf{Environment Context}}}, colframe=systemprompt, coltitle=white, colbacktitle=systemprompt]

\textbf{Scene Setup:}
\begin{itemize}
    \item Robotic arm on table
    \item Laptop on table
    \item Human standing next to table
    \item Task: Learn to manipulate a cup based on language commands
\end{itemize}

\vspace{3mm}

\textbf{State Space (19 dimensions):}

\begin{tabular}{ll}
\textcolor{instruction}{\textbf{Robot End Effector (12 dims):}} & \\
\quad Positions (3): & $x, y, z$ \\
\quad Rotation matrix (9): & $R_{xx}, R_{xy}, R_{xz},$ \\
\quad &$R_{yx}, R_{yy}, R_{yz},$ \\
\quad &$R_{zx}, R_{zy}, R_{zz}$ \\
\textcolor{instruction}{\textbf{Environment Objects (7 dims):}} & \\
\quad Human (3): & $x, y, z$ \\
\quad Laptop (3): & $x, y, z$ \\
\quad Table (1): & $z$ \\
\end{tabular}

\end{tcolorbox}

\begin{tcolorbox}[promptbox, title={\textcolor{white}{\textbf{Instruction Input}}}, colframe=userprompt, coltitle=white, colbacktitle=userprompt]

\textbf{Language Instruction:} \textcolor{variable}{\texttt{[instruction]}}

\textit{Example: "Stay close to the table surface. Carry the cup upright."}

\end{tcolorbox}

\begin{tcolorbox}[promptbox, title={\textcolor{white}{\textbf{Distance Reasoning Guidelines}}}, colframe=instruction, coltitle=white, colbacktitle=instruction]

\textbf{General Principles:}
\begin{itemize}
    \item Consider specific axes/planes depending on instruction context
    \item For bulky objects (e.g., human): use horizontal distance in $xy$-plane
\end{itemize}

\vspace{3mm}

\textbf{Orientation/Direction Reasoning:}
\begin{itemize}
    \item End effector's $x$-axis points \textbf{upward} from grasped object
    \item PyBullet global "up" direction = world $z$-axis
    \item Rotation matrix element $R_{ji}$ = alignment between local axis $i$ and global axis $j$
    \item To align end effector's axis to world axis: identify corresponding rotation matrix element
\end{itemize}

\end{tcolorbox}

\begin{tcolorbox}[promptbox, title={\textcolor{white}{\textbf{Task Instructions}}}, colframe=variable, coltitle=white, colbacktitle=variable]

\textbf{Step-by-step reasoning required:}
\begin{enumerate}
    \item For each of the 19 state elements, explain whether robot needs to pay attention to it
    \item Consider instruction requirements carefully
\end{enumerate}

\vspace{3mm}

\textbf{Output Format:} JSON object with binary arrays (0 = ignore, 1 = attend)

\begin{verbatim}
{
  "eef_pos": [d1, d2, d3],
  "eef_rot": [d4, ..., d12],
  "human": [d13, d14, d15],
  "laptop": [d16, d17, d18],
  "table": [d19]
}
\end{verbatim}

\textit{Output this on a new line with no additional text.}

\end{tcolorbox}

\subsection{Training Details}

We used a single L40 NVIDIA GPU to train each model. We used a learning rate of $1e-3$, and the reward model MLP consists of three hidden layers with hidden sizes $128$, $256$, and $128$. We used batch size $512$ for training. $\lambda$ is chosen as $10$ and $1$ for simulation and real robot experiments, respectively.

\end{document}